\documentclass{article} % For LaTeX2e
\usepackage[ruled,noline,commentsnumbered]{algorithm2e}
\usepackage{nips14submit_e,times}
\usepackage{hyperref}
\usepackage{url}
\usepackage{amssymb,amsmath,etoolbox}
\usepackage{siunitx}
\usepackage{amsfonts}
\usepackage{fixmath}
\usepackage{bm}
\usepackage{enumitem}
\usepackage{amsmath}

\title{A Boosting Framework on Grounds of Online Learning}

\author{
Tofigh ~Naghibi, Beat ~Pfister \\
Computer Engineering and Networks Laboratory\\
ETH Zurich, Switzerland \\
\texttt{naghibi@tik.ee.ethz.ch, pfister@tik.ee.ethz.ch} 
}

\makeatletter
\newif\if@gather@prefix 
\preto\place@tag@gather{% 
  \if@gather@prefix\iftagsleft@ 
    \kern-\gdisplaywidth@ 
    \rlap{\gather@prefix}% 
    \kern\gdisplaywidth@ 
  \fi\fi 
} 
\appto\place@tag@gather{% 
  \if@gather@prefix\iftagsleft@\else 
    \kern-\displaywidth 
    \rlap{\gather@prefix}% 
    \kern\displaywidth 
  \fi\fi 
  \global\@gather@prefixfalse 
} 
\preto\place@tag{% 
  \if@gather@prefix\iftagsleft@ 
    \kern-\gdisplaywidth@ 
    \rlap{\gather@prefix}% 
    \kern\displaywidth@ 
  \fi\fi 
} 
\appto\place@tag{% 
  \if@gather@prefix\iftagsleft@\else 
    \kern-\displaywidth 
    \rlap{\gather@prefix}% 
    \kern\displaywidth 
  \fi\fi 
  \global\@gather@prefixfalse 
} 
\def\math@cr@@@align{%
  \ifst@rred\nonumber\fi
  \if@eqnsw \global\tag@true \fi
  \global\advance\row@\@ne
  \add@amps\maxfields@
  \omit
  \kern-\alignsep@
  \if@gather@prefix\tag@true\fi
  \iftag@
    \setboxz@h{\@lign\strut@{\make@display@tag}}%
    \place@tag
  \fi
  \ifst@rred\else\global\@eqnswtrue\fi
  \global\lineht@\z@
  \cr
}
\newcommand*{\beforetext}[1]{% 
  \ifmeasuring@\else
  \gdef\gather@prefix{#1}% 
  \global\@gather@prefixtrue 
  \fi
} 
\makeatother
\DeclareMathOperator*{\quadr}{\mathcal{R}(\mathbf{w})\!=\frac{1}{2}||\mathbf{w}||_2^2}

\DeclareMathOperator*{\argmin}{arg\,min}
\newcommand{\eref}[1]{(\ref{#1})}
\DeclareMathSymbol{\Pi}{\mathalpha}{operators}{5}
\nipsfinalcopy % Uncomment for camera-ready version

\begin{document}

\maketitle
\begin{abstract}
 By exploiting the duality between boosting and online learning, we present a boosting framework which proves to be extremely powerful thanks to employing the vast knowledge available in the online learning area. Using this framework, we develop various algorithms to address multiple practically and theoretically interesting questions including sparse boosting, smooth-distribution boosting, agnostic learning and, as a by-product, some generalization to double-projection online learning algorithms\footnote{Boosting algorithms in this paper can be found in `maboost' R package}.
\end{abstract}

\section{Introduction}
A boosting algorithm can be seen as a meta-algorithm that maintains a distribution over the sample space. At each iteration a weak hypothesis is learned and the distribution is updated, accordingly. The output (strong hypothesis) is a convex combination of the weak hypotheses. Two dominant views to describe and design boosting algorithms are ``weak to strong learner'' (WTSL), which is the original viewpoint presented in \cite{schapire:90,freund:97}, and boosting by ``coordinate-wise gradient descent in the functional space'' (CWGD) appearing in later works \cite{breiman:99,mason:99,friedman:00}. A boosting algorithm adhering to the first view guarantees that it only requires a finite number of iterations (equivalently, finite number of weak hypotheses) to learn a $(1\!-\epsilon)$-accurate hypothesis. In contrast, an algorithm resulting from the CWGD viewpoint (usually called potential booster) may not necessarily be a boosting algorithm in the probability approximately correct (PAC) learning sense. However, 
while it is rather difficult to 
construct a boosting algorithm based on the 
first view, the algorithmic frameworks, e.g., AnyBoost \cite{mason:99}, resulting from the second viewpoint have proven to be particularly prolific when it comes to developing new boosting algorithms. Under the CWGD view, the choice of the convex loss function to be minimized is (arguably) the cornerstone of designing a boosting algorithm. This, however, is a severe disadvantage in some applications.

In CWGD, the weights are not directly controllable (designable) and are only viewed as the values of the gradient of the loss function. In many applications, some characteristics of the desired distribution are known or given as problem requirements while, finding a loss function that generates such a distribution is likely to be difficult. For instance, what loss functions can generate sparse distributions?\footnote{In the boosting terminology, sparsity usually refers to the greedy hypothesis-selection strategy of boosting methods in the functional space. However, sparsity in this paper refers to the sparsity of the distribution (weights) over the sample space.} What family of loss functions results in a smooth distribution?\footnote{A smooth distribution is a distribution that does not put too much weight on any single sample or in other words, a distribution emulated by the 
booster does not dramatically diverge from the target distribution \cite{servedio:03, gavinsky:03}.} We even can go further and imagine the scenarios in which a loss function needs to put more weights on a given subset of examples than others,  either because that subset has more reliable labels or it is a problem requirement to have a more accurate hypothesis for that part of the sample space. Then, what loss function can generate such a customized distribution? Moreover, does it result in a provable boosting algorithm? In general, how can we characterize the accuracy of the final hypothesis?

Although, to be fair, the so-called loss function hunting approach has given rise to useful boosting algorithms such as LogitBoost, FilterBoost, GiniBoost and MadaBoost \cite{friedman:00,bradley:08,hatano:06,domingo:00} which (to some extent) answer some of the above questions, it is an inflexible and relatively unsuccessful approach to addressing the boosting problems with distribution constraints.

Another approach to designing a boosting algorithm is to directly follow the WTSL viewpoint \cite{freund:95,servedio:03,bshouty:02}. The immediate advantages of such an approach are, first, the resultant algorithms are provable boosting algorithms, i.e., they output a hypothesis of arbitrary accuracy. Second, the booster has direct control over the weights, making it more suitable for boosting problems 
subject to some distribution constraints. However, since the WTSL view does not offer any algorithmic framework (as opposed to the CWGD view), it is rather difficult to come up with a distribution update mechanism resulting in a provable boosting algorithm. There are, however, a few useful, and albeit fairly limited, algorithmic frameworks such as TotalBoost \cite{warmuth:06} that can be used to derive other provable boosting algorithms. The TotalBoost algorithm can maximize the margin by iteratively solving a convex problem with the totally corrective constraint. A more general family of boosting algorithms was later proposed by Shalev-Shwartz et. al. \cite{shwartz:08}, where it was shown that weak learnability and linear separability are equivalent, a result following from von Neumann's minmax theorem.
 Using this theorem, they constructed a family of algorithms that maintain smooth distributions over the sample space, and consequently are noise tolerant. Their proposed algorithms find an $(1\!-\epsilon)$-accurate solution after performing at most $O(\log(N)/\epsilon^2)$ iterations, where $N$ is the number of training examples. 
\subsection{Our Results}
We present a family of boosting algorithms that can be derived from well-known online learning algorithms, including projected gradient descent \cite{zinkevich:03} and its generalization, mirror descent (both active and lazy updates, see \cite{hazan:09}) and composite objective mirror descent (COMID) \cite{duch:10}. We prove the PAC learnability of the algorithms derived from this framework and we show that this framework in fact generates maximum margin algorithms. That is, given a desired accuracy level $\nu$, it outputs a hypothesis of margin $\gamma_{\text{min}}-\nu$ with $\gamma_{\text{min}}$ being the minimum edge that the weak classifier guarantees to return.

The duality between (linear) online learning and boosting is by no means new. This duality was first pointed out in \cite{freund:97} and was later elaborated and formalized by using the von Neumann's minmax theorem \cite{freund96}. Following this line, we provide several proof techniques required to show the PAC learnability of the derived boosting algorithms. These techniques are fairly versatile and can be used to translate many other online learning methods into our boosting framework. To motivate our boosting framework, we derive two practically and theoretically interesting algorithms: (I) SparseBoost algorithm which by maintaining a sparse distribution over the sample space tries to reduce the space and the computation complexity. In fact this problem, i.e., applying batch boosting on the successive subsets of data when there is not sufficient memory to store an entire dataset, was first discussed by Breiman in \cite{breiman:97}, though no algorithm with theoretical guarantee was suggested. SparseBoost 
is the first provable batch booster that can (partially) address this problem. By analyzing this algorithm, we show that the tuning parameter of the regularization term $\ell_1$ at each round $t$ should not exceed $\frac{\gamma_t}{2}\eta_t$ to still have a boosting algorithm, where $\eta_t$ is the coefficient of the $t^{\text{th}}$ weak hypothesis and $\gamma_t$ is its edge. (II) A smooth boosting algorithm that requires only $O(\log 1/\epsilon)$ number of rounds to learn a $(1\!-\epsilon)$-accurate hypothesis. This algorithm can also be seen as an agnostic boosting 
algorithm\footnote{Unlike the PAC model, the agnostic learning model allows an arbitrary target function (labeling function) that may not belong to the class studied, and hence, can be viewed as a noise tolerant learning model \cite{kearns:92}.} due to the fact that smooth distributions provide a theoretical guarantee for noise tolerance in various noisy learning settings, such as agnostic boosting \cite{kalai:09,ben:01}.  

Furthermore, we provide an interesting theoretical result about MadaBoost \cite{domingo:00}. We give a proof (to the best of our knowledge the only available unconditional proof) for the boosting property of (a variant of) MadaBoost and show that, unlike the common presumption, its convergence rate is of $O(1/\epsilon^2)$ rather than $O(1/\epsilon)$.
 
Finally, we show our proof technique can be employed to generalize some of the known online learning algorithms. Specifically, consider the Lazy update variant of the online Mirror Descent (LMD) algorithm (see for instance \cite{hazan:09}). The standard proof to show that the LMD update scheme achieves vanishing regret bound is through showing its equivalence to the FTRL algorithm \cite{hazan:09} in the case that they are both linearized, i.e., the cost function is linear. However, this indirect proof is fairly restrictive when it comes to generalizing the LMD-type algorithms. Here, we present a direct proof for it, which can be easily adopted to generalize the LMD-type algorithms. 
\vspace{-2mm}
\section{Preliminaries}
Let $\{(\mathbf{x}_i,a_i)\}, 1\leq i \leq N$, be $N$ training samples, where $\mathbf{x}_i\!\in\mathcal{X}$ and $a_i\!\in\{-1,+1\}$. Assume $h\in \mathcal{H}$ is a real-valued function mapping $\mathcal{X}$ into $[-1,1]$. Denote a   distribution over the training data by $\mathbf{w}=[w_1,\dots,w_N]^{\top}$ and define a loss vector $\mathbf{d}=[-a_1h(\mathbf{x}_1),\dots,-a_Nh(\mathbf{x}_N)]^{\top}$. We define $\gamma= -\mathbf{w}^{\top}\mathbf{d}$ as the \textit{edge} of the hypothesis $h$ under the distribution $w$ and it is assumed to be positive when $h$ is returned by a weak learner. In this paper we do not consider the branching program based boosters and adhere to the typical boosting protocol (described in Section 1).  

Since a central notion throughout this paper is that of Bregman divergences, we briefly revisit some of their properties. A Bregman divergence is defined with respect to a convex function $\mathcal{R}$ as
\begin{align} 
\label{eq_2}
B_{\mathcal{R}}(\mathbf{x},\mathbf{y}) \!=  \mathcal{R}(\mathbf{x})-\mathcal{R}(\mathbf{y})- \nabla \mathcal{R}(\mathbf{y})(\mathbf{x}-\mathbf{y})^{\top}
\end{align}
and can be interpreted as a distance measure between $\mathbf{x}$ and $\mathbf{y}$. Due to the convexity of $\mathcal{R}$, a Bregman divergence is always non-negative, i.e., $B_{\mathcal{R}}(\mathbf{x},\mathbf{y})\ge0 $. In this work we consider $\mathcal{R}$ to be a $\beta$-strongly convex function\footnote{That is, its second derivative (Hessian in higher dimensions) is bounded away from zero by at least $\beta$.} with respect to a norm $||.||$. With this choice of $\mathcal{R}$, the Bregman divergence $B_{\mathcal{R}}(\mathbf{x},\mathbf{y})\ge \frac{\beta}{2}||\mathbf{x}-\mathbf{y} ||^2$. As an example, if $\mathcal{R}(\mathbf{x})=\frac{1}{2}\mathbf{x}^{\top}\mathbf{x}$ (which is 1-strongly convex with respect to $||.||_2$), then $B_{\mathcal{R}}(\mathbf{x},\mathbf{y})=\frac{1}{2}||\mathbf{x}-\mathbf{y}||_2^2$ is the Euclidean distance. 
Another example is the negative entropy function $\mathcal{R}(\mathbf{x})=\sum_{i=1}^N x_i\log x_i$  (resulting in the KL-divergence) which is known to be 1-strongly convex over the probability simplex with respect to $\ell_1$ norm.   

The Bregman projection is another fundamental concept of our framework. 

\textbf{Definition 1 (Bregman Projection).} \textit{The Bregman projection of a vector $\mathbf{y}$ onto a convex set $\mathcal{S}$ with respect to a Bregman divergence $B_{\mathcal{R}}$ is}
\begin{align} 
\label{eq_4}
\Pi_{\mathcal{S}}(\mathbf{y}) = \argmin_{\mathbf{x}\in \mathcal{S}} B_{\mathcal{R}}(\mathbf{x},\mathbf{y})
\end{align}
Moreover, the following generalized Pythagorean theorem holds for Bregman projections.

\textbf{Lemma 1 (Generalized Pythagorean)} \cite[Lemma 11.3]{bianchi:06}\textbf{.} \textit{Given a point $\mathbf{y}\!\in \mathbb{R}^N$, a convex set $\mathcal{S}$ and $\hat{\mathbf{y}}\!=\Pi_{\mathcal{S}}(\mathbf{y})$ as the Bregman projection of $\mathbf{y}$ onto $\mathcal{S}$, for all $\mathbf{x}\in \mathcal{S}$ we have}
\begin{align} 
\label{eq_5}
\text{Exact:}& \quad\quad B_{\mathcal{R}}(\mathbf{x},\mathbf{y})\ge B_{\mathcal{R}}(\mathbf{x},\hat{\mathbf{y}})+ B_{\mathcal{R}}(\hat{\mathbf{y}},\mathbf{y})\\
\text{Relaxed:}& \quad\quad B_{\mathcal{R}}(\mathbf{x},\mathbf{y})\ge B_{\mathcal{R}}(\mathbf{x},\hat{\mathbf{y}})
\end{align}
The relaxed version follows from the fact that $B_{\mathcal{R}}(\hat{\mathbf{y}},\mathbf{y})\!\ge\!0$ and thus can be ignored.

\textbf{Lemma 2.} \textit{For any vectors $\mathbf{x},\mathbf{y},\mathbf{z}$, we have}
\begin{align} 
\label{eq_6}
(\mathbf{x}-\mathbf{y})^{\top}(\nabla\mathcal{R}(\mathbf{z})-\nabla\mathcal{R}(\mathbf{y}))=B_{\mathcal{R}}(\mathbf{x},\mathbf{y})-B_{\mathcal{R}}(\mathbf{x},\mathbf{z})+B_{\mathcal{R}}(\mathbf{y},\mathbf{z})
\end{align}
The above lemma follows directly from the Bregman divergence definition in \eref{eq_2}. Additionally, the following definitions from convex analysis are useful throughout the paper.

\textbf{Definition 2 (Norm \& dual norm).} \textit{Let $||.||_A$ be a norm. Then its dual norm is defined as}
\begin{align} 
\label{eq_7}
&||\mathbf{y}||_{A^*} = \sup \{\mathbf{y}^\top\mathbf{x}, ||\mathbf{x}||_A\le1\}
\end{align}
For instance, the dual norm of $||.||_2=\ell_2$ is $||.||_{2^*}=\ell_2$ norm and the dual norm of $\ell_1$ is $\ell_\infty$ norm. Further, 

\textbf{Lemma 3.} \textit{For any vectors $\mathbf{x},\mathbf{y}$ and any norm $||.||_A$, the following inequality holds:}
\begin{align} 
\label{eq_8}
\mathbf{x}^{\top}\mathbf{y} \le ||\mathbf{x}||_A ||\mathbf{y}||_{A^*}\le \frac{1}{2}||\mathbf{x}||^2_A+ \frac{1}{2}||\mathbf{y}||_{A^*}^{2}
\end{align}
Throughout this paper, we use the shorthands $||.||_{A}\!= ||.||$ and $||.||_{A^*}\!= ||.||_*$ for the norm and its dual, respectively.

Finally, before continuing, we establish our notations. Vectors are lower case bold letters and their entries are non-bold letters with subscripts, such as $x_i$ of $\mathbf{x}$, or non-bold letter with superscripts if the vector already has a subscript, such as $x_t^i$ of $\mathbf{x}_t$. Moreover, an N-dimensional probability simplex is denoted by $\mathcal{S}=\{\mathbf{w}|\sum_{i=1}^N w_i=1 , w_i\ge0\}$. The proofs of the theorems and the lemmas can be found in the Supplement.
\vspace{-2mm}
\section{Boosting Framework}
\vspace{-2mm}

Let $\mathcal{R}(\mathbf{x})$ be a $1$-strongly convex function with respect to a norm $||.||$ and denote its associated Bregman divergence $B_{\mathcal{R}}$. Moreover, let the dual norm of a loss vector $\mathbf{d}_t$ be upper bounded, i.e., $||\mathbf{d}_t||_*\le L$. It is easy to verify that for $\mathbf{d}_t$ as defined in MABoost, $L\!=1$ when $||.||_*=\ell_\infty$ and $L\!=N$ when $||.||_*=\ell_2$. The following Mirror Ascent Boosting (MABoost) algorithm is our boosting framework.
\vspace{2mm}
\begin{algorithm}[H]
%\SetKw{Initialize}{Initialize}
\SetKwInput{Input}{Input}
\SetKwInOut{Output}{Output}
  \caption{Mirror Ascent Boosting (MABoost)}
  \begin{tabbing}
  \hspace{8mm}\=\hspace{7mm}\=\hspace{30mm}\=\hspace{50mm}\= \kill
  \textbf{Input:} ~$\mathcal{R}(\mathbf{x})$ $1$-strongly convex function, $\mathbf{w}_1=[\frac{1}{N},\dots,\frac{1}{N}]^{\top}$ and $\mathbf{z}_1=[\frac{1}{N},\dots,\frac{1}{N}]^{\top}$ \\[2mm]
  \textbf{For} ~{$t=1,\dots,T$} \textbf{do} \\[1mm]
  \>(a) \>Train classifier with $\mathbf{w}_t$ and get $h_t$, let $\mathbf{d}_{t}=[-a_1h_{t}(\mathbf{x}_1),\dots,-a_Nh_{t}(\mathbf{x}_N)]$ \\
  \>    \>and $\gamma_t= -\mathbf{w}_t^{\top}\mathbf{d}_t$. \\[2mm]
  \>(b) \>Set $\eta_t=\frac{\gamma_t}{L}$ \\[2mm]
  \>(c) \>Update weights: \>$\nabla\mathcal{R}(\mathbf{z}_{t+1}) = \nabla\mathcal{R}(\mathbf{z}_{t})+\eta_t\mathbf{d}_{t}$ \>(lazy update) \\[1mm]
  \>    \>              \>$\nabla\mathcal{R}(\mathbf{z}_{t+1}) =  \nabla\mathcal{R}(\mathbf{w}_{t})+\eta_t\mathbf{d}_{t}$ \>(active update) \\[2mm]
  \>(d) \>Project onto $\mathcal{S}$: \> 
                    $\mathbf{w}_{t+1}=\underset{\mathbf{w}\in \mathcal{S}}{\operatorname{argmin}}\, B_{\mathcal{R}}(\mathbf{w},\mathbf{z}_{t+1})$  \\[1mm]
                    
  \textbf{End} \\
  \textbf{Output:} ~The final hypothesis $f(\mathbf{x}) \!=\text{sign}\bigg(\sum_{t=1}^T \eta_t h_{t}(\mathbf{x})\bigg)$.
  \end{tabbing}
\end{algorithm}
\vspace{1mm}
%$\mathbf{w}_{t+1}=\argmin_{\mathbf{w}\in \mathcal{S}} B_{\mathcal{R}}(\mathbf{w},\mathbf{z}_{t+1})$
 This algorithm is a variant of the mirror descent algorithm \cite{hazan:09}, modified to work as a boosting algorithm. The basic principle in this algorithm is quite clear. As in ADABoost, the weight of a wrongly (correctly) classified sample increases (decreases). The weight vector is then projected onto the probability simplex in order to keep the weight sum equal to 1. The distinction between the active and lazy update versions and the fact that the algorithm may behave quite differently under different update strategies should be emphasized. In the lazy update version, the norm of the auxiliary variable $\mathbf{z}_t$ is unbounded which makes the lazy update inappropriate in some situations. In the active update version, on the other hand, the algorithm always needs to access (compute) the previous projected weight $\mathbf{w}_{t}$ to update the weight at round $t$ and this may not be possible in some applications (such as boosting-by-filtering). 
 
Due to the duality between online learning and boosting, it is not surprising that MABoost (both the active and lazy versions) is a boosting algorithm. The proof of its boosting property, however, reveals some interesting properties  which enables us to generalize the MABoost framework. In the following, only the proof of the active update is given and the lazy update is left to Section 3.4.

\textbf{Theorem 1.} \textit{Suppose that MABoost generates weak hypotheses $h_1,\dots,h_T$ whose edges are $\gamma_1,\dots,\gamma_T$. Then the error $\epsilon$ of the combined hypothesis $f$ on the training set is bounded as}:
\begin{align} 
\mathcal{R}(\mathbf{w})=\frac{1}{2}||\mathbf{w}||_2^2: \quad\quad\quad\quad\quad\quad&\epsilon \le \frac{1}{1+\sum_{t=1}^T \gamma_t^2 } \label{eq_10}\\
\mathcal{R}(\mathbf{w})\!=\sum_{i=1}^{N} w_i\log{w_i}: \quad\quad\quad\quad\quad\quad    & \epsilon \le e^{-\sum_{t=1}^T \frac{1}{2} \gamma_t^2}   \label{eq_11} 
\end{align}
 In fact, the first bound \eref{eq_10} holds for any 1-strongly convex $\mathcal{R}$, though for some $\mathcal{R}$ (e.g., negative entropy) a much tighter bound as in \eref{eq_11} can be achieved.
 
 \textit{Proof}: Assume $\mathbf{w}^*=[w_1^*,\dots\ ,w_N^*]^{\top}$ is a distribution vector where $w^*_i=\frac{1}{N\epsilon}$ if $f(\mathbf{x}_i)\neq a_i$, and 0 otherwise. $\mathbf{w}^*$ can be seen as a uniform distribution over the wrongly classified samples by the ensemble hypothesis $f$. Using this vector and following the approach in \cite{hazan:09}, we derive the upper bound of $\sum_{t=1}^T \eta_t(\mathbf{w}^{*\top}\mathbf{d}_t  {-}\mathbf{w}_t^{\top}\mathbf{d}_t)$ where $\mathbf{d}_t=[d_t^1,\dots,\!d_t^N]$ is a loss vector as defined in Algorithm~1. 
\begin{subequations}
\label{equations}
\begin{align}
  \label{eq:f}
 (\mathbf{w}^*\!-\mathbf{w}_{t})^{\top}\eta_t\mathbf{d}_{t}\!&=(\mathbf{w}^*-\mathbf{w}_{t})^{\top}\big(\nabla\mathcal{R}(\mathbf{z}_{t+1})-\nabla\mathcal{R}(\mathbf{w}_{t})\big)\\
  \label{eq:g}
 &=B_{\mathcal{R}}(\mathbf{w}^*,\mathbf{w}_{t})  -  B_{\mathcal{R}}(\mathbf{w}^* , \mathbf{z}_{t+1}) +  B_{\mathcal{R}}(\mathbf{w}_t,\mathbf{z}_{t+1})  \\
  \label{eq:h}
 &\le  B_{\mathcal{R}}(\mathbf{w}^*,\mathbf{w}_{t}) -   B_{\mathcal{R}}(\mathbf{w}^* , \mathbf{w}_{t+1}) + B_{\mathcal{R}}(\mathbf{w}_t,\mathbf{z}_{t+1})  
 \end{align}
\end{subequations}
where the first equation follows Lemma 2 and inequality \eref{eq:h} results from the relaxed version of Lemma 1. Note that Lemma 1 can be applied here because $\mathbf{w}^*\!\in\mathcal{S}$.

Further, the $B_{\mathcal{R}}(\mathbf{w}_t,\mathbf{z}_{t+1})$ term is bounded. By applying Lemma 3
 \begin{align} 
\label{eq_12.5}
B_{\mathcal{R}}(\mathbf{w}_t,\mathbf{z}_{t+1}) + B_{\mathcal{R}}(\mathbf{z}_{t+1},\mathbf{w}_t)= (\mathbf{z}_{t+1}-\mathbf{w}_t)^\top \eta_t\mathbf{d}_t\le \frac{1}{2}||\mathbf{z}_{t+1}-\mathbf{w}_t||^2 +\frac{ 1 }{2}\eta_t^2||\mathbf{d}_t||_*^2
\end{align}
and since $B_{\mathcal{R}}(\mathbf{z}_{t+1},\mathbf{w}_t)\ge \frac{1}{2}||\mathbf{z}_{t+1}-\mathbf{w}_t||^2$ due to the $1$-strongly convexity of $\mathcal{R}$, we have
 \begin{align} 
\label{eq_12.8}
B_{\mathcal{R}}(\mathbf{w}_t,\mathbf{z}_{t+1})\le  \frac{1}{2}\eta_t^2||\mathbf{d}_t||_*^2
\end{align}
 Now, replacing \eref{eq_12.8} into \eref{eq:h} and summing it up from $t=1$ to $T$, yields 
   \begin{align} 
\label{eq_13}
 \sum_{t=1}^{T} \mathbf{w}^{*\top}\eta_t\mathbf{d}_{t}\!-\mathbf{w}_{t}^{\top}\eta_t\mathbf{d}_{t}\le \sum_{t=1}^T \frac{1}{2}\eta_t^2||\mathbf{d}_{t}||^2_*  +   B_{\mathcal{R}}(\mathbf{w}^*,\mathbf{w}_{1}) -  B_{\mathcal{R}}(\mathbf{w}^*,\mathbf{w}_{T+1}) 
\end{align}
 Moreover, it is evident from the algorithm description that for mistakenly classified samples
 \begin{align}
 \label{eq_14}
  -a_i f(\mathbf{x}_i) \!=-a_i\text{sign}\bigg(\sum_{t=1}^T \eta_t h_{t}(\mathbf{x}_i)\bigg)=\text{sign}\bigg(\sum_{t=1}^T \eta_t d_t^i\bigg)\ge0  \quad\forall \mathbf{x}_i\in \{\mathbf{x} | f(\mathbf{x}_i)\neq a_i\} 
\end{align}  
 Following \eref{eq_14}, the first term in \eref{eq_13} will be $\mathbf{w}^{*\top}\sum_{t=1}^T \eta_t \mathbf{d}_t\!\ge0$ and thus, can be ignored. Moreover, by the definition of $\gamma$, the second term is $\sum_{t=1}^T -\mathbf{w}_t^{\top}\eta_t\mathbf{d}_t\!=\sum_{t=1}^T \eta_t\gamma_t$. Putting all these together, ignoring the last term in \eref{eq_13} and replacing $||\mathbf{d}_{t}||^2_*$ with its upper bound $L$, yields
\begin{align} 
\label{eq_15}
  -B_{\mathcal{R}}(\mathbf{w}^*,\mathbf{w}_{1})\le L\sum_{t=1}^T \frac{1}{2}\eta_t^2  - \sum_{t=1}^T \eta_t\gamma_t 
\end{align}
Replacing the left side with $-B_{\mathcal{R}}\!=-\frac{1}{2}||\mathbf{w}^*\!-\mathbf{w}_{1}||^2\!=\frac{\epsilon-1}{2N\epsilon}$ for the case of quadratic $\mathcal{R}$, and with $-B_{\mathcal{R}}\!=\log(\epsilon)$ when $\mathcal{R}$ is a negative entropy function, taking the derivative w.r.t $\eta_t$ and equating it to zero (which yields $\eta_t=\frac{\gamma_t}{L}$) we achieve the error bounds in \eref{eq_10} and \eref{eq_11}. Note that in the case of $\mathcal{R}$ being the negative entropy function, Algorithm 1 degenerates into ADABoost with a different choice of $\eta_t$.  

Before continuing our discussion, it is important to mention that the cornerstone concept of the proof is the choice of $\mathbf{w}^*$. For instance, a different choice of $\mathbf{w}^*$ results in the following max-margin theorem.

\textbf{Theorem 2.} \textit{Setting $\eta_t=\frac{\gamma_t}{L\sqrt {t}}$, MABoost outputs a hypothesis of margin at least $\gamma_\text{min}-\nu$, where $\nu$ is a desired accuracy level and tends to zero in $O(\frac{\log T}{\sqrt{T}})$ rounds of boosting}.

 \textbf{Observations:} Two observations follow immediately from the proof of Theorem 1. First, the requirement of using Lemma 1 is $\mathbf{w}^*\!\in\!\mathcal{S}$, so in the case of projecting onto a smaller convex set $\mathcal{S}_k\!\subseteq\!\mathcal{S}$, as long as $\mathbf{w}^*\!\in\!\mathcal{S}_k$ holds, the proof is intact. Second, only the relaxed version of Lemma~1 is required in the proof (to obtain inequality \eref{eq:h}). Hence, if there is an approximate projection operator $\hat{\Pi}_{\mathcal{S}}$ that satisfies the inequality $B_{\mathcal{R}}(\mathbf{w}^*,\mathbf{z}_{t+1})\ge B_{\mathcal{R}}\big(\mathbf{w}^*,\hat{\Pi}_{\mathcal{S}}(\mathbf{z}_{t+1})\big)$, it can be substituted for the exact projection operator $\Pi_{\mathcal{S}}$ and the active update version of the algorithm still works. A practical approximate operator of this type can be obtained through a double-projection strategy.
 
 \textbf{Lemma 4.} \textit{Consider the convex sets $\mathcal{K}$ and $\mathcal{S}$, where $\mathcal{S}\!\subseteq\!\mathcal{K}$. Then for any $\mathbf{x}\!\in\!\mathcal{S}$ and $\mathbf{y}\!\in\!\mathbb{R}^N$, $\hat{\Pi}_{\mathcal{S}}(\mathbf{y})\!=\!\Pi_{\mathcal{S}}\Big(\Pi_{\mathcal{K}}(\mathbf{y})\Big)$ is an approximate projection that satisfies $B_{\mathcal{R}}(\mathbf{x},\mathbf{y})\!\ge \!B_{\mathcal{R}}\big(\mathbf{x},\hat{\Pi}_{\mathcal{S}}(\mathbf{y})\big)$}.
 
These observations are employed to generalize Algorithm 1. However, we want to emphasis that the approximate Bregman projection is only valid for the active update version of MABoost.
\vspace{-2mm}
\subsection{Smooth Boosting}
\vspace{-2mm}
Let $k\!>0$ be a smoothness parameter. A distribution $\mathbf{w}$ is smooth w.r.t a given distribution $\mathbf{D}$ if $w_i\!\le kD_i$ for all $1\!\le i\!\le N$. Here, we consider the smoothness w.r.t to the uniform distribution, i.e., $D_i\!=\frac{1}{N}$. Then, given a desired smoothness parameter $k$, we require a boosting algorithm that only constructs distributions $\mathbf{w}$ such that $w_i\!\le \frac{k}{N}$, while guaranteeing to output a $(1\!-\frac{1}{k})$-accurate hypothesis. To this end, we only need to replace the probability simplex $\mathcal{S}$ with $\mathcal{S}_k\!=\{\mathbf{w}|\sum_{i=1}^N w_i\!=1 , 0\!\le w_i\!\le \frac{k}{N}\}$ in MABoost to obtain a smooth distribution boosting algorithm, called smooth-MABoost.  That is, the update rule is: $ \mathbf{w}_{t+1}=\underset{\mathbf{w}\in \mathcal{S}_k} {\mathrm{argmin}} ~B_{\mathcal{R}}(\mathbf{w},\mathbf{z}_{t+1})$.

Note that the proof of Theorem 1 holds for smooth-MABoost, as well. As long as $\epsilon\!\ge\frac{1}{k}$, the error distribution $\mathbf{w}^*$ ($w^*_i\!=\frac{1}{N\epsilon}$ if $f(\mathbf{x}_i)\neq a_i$, and 0 otherwise) is in $\mathcal{S}_k$ because $\frac{1}{N\epsilon}\le\frac{k}{N}$. Thus, based on the first observation, the error bounds achieved in Theorem 1 hold for $\epsilon\!\ge\frac{1}{k}$. In particular, $\epsilon\!=\frac{1}{k}$ is reached after a finite number of iterations. This projection problem has already appeared in the literature. An entropic projection algorithm ($\mathcal{R}$ is negative entropy), for instance, was proposed in \cite{shwartz:08}. Using negative entropy and their suggested projection algorithm results in a fast smooth boosting algorithm with the following convergence rate.

\textbf{Theorem 3.} \textit{Given $\mathcal{R}(\mathbf{w})\!=\sum_{i=1}^{N} w_i\log{w_i}$ and a desired $\epsilon$, smooth-MABoost finds a $(1-\epsilon)$-accurate hypothesis in $O(\log(\frac{1}{\epsilon})/\gamma^2)$ of iterations.}
\vspace{-2mm}
\subsection{Combining Datasets}
 Let's assume we have two sets of data. A primary dataset $\mathcal{A}$ and a secondary dataset $\mathcal{B}$. The goal is to train a classifier that achieves $(1\!-\epsilon)$ accuracy on $\mathcal{A}$ while limiting the error on dataset  $\mathcal{B}$ to $\epsilon_{\mathcal{B}}\le\frac{1}{k}$. This scenario has many potential applications including transfer learning \cite{dai:07}, weighted combination of datasets based on their noise level and emphasizing on a particular region of a sample space as a problem requirement (e.g., a medical diagnostic test that should not make a wrong diagnosis when the sample is a pregnant woman). To address this problem, we only need to replace  $\mathcal{S}$ in MABoost with $\mathcal{S}_c\!=\{\mathbf{w}|\sum_{i=1}^N w_i\!=1 , 0\!\le w_i\enskip \forall i\in\mathcal{A} \enskip \wedge \enskip 0\!\le w_i\!\le \frac{k}{N}\enskip \forall i\in\mathcal{B}  \}$ where $i\in\mathcal{A}$ shorthands the indices of samples in $\mathcal{A}$. By generating smooth distributions on $\mathcal{
B}$, 
this algorithm limits the weight of the secondary dataset, which intuitively results in limiting its effect on the final hypothesis. The proof of its boosting property is quite similar to Theorem 1 (see supplement).
\vspace{-2mm}
\subsection{Sparse Boosting}
  Let $\quadr$. Since in this case the projection onto the simplex is in fact an $\ell_1$-constrained optimization problem, it is plausible that some of the weights are zero (sparse distribution), which is already a useful observation. To promote the sparsity of the weight vector, we want to directly regularize the projection with the $\ell_1$ norm, i.e., adding $||\mathbf{w}||_1$ to the objective function in the projection step. It is, however, not possible in MABoost, since $||\mathbf{w}||_1$ is trivially constant on the simplex. Therefore, we split the projection step into two consecutive steps. The first projection is onto $\mathcal{R}_{+}\!=\!\{\mathbf{y}\enspace|\enspace0\!\le y_i\}$.
  
  Surprisingly, projection onto $\mathcal{R}_{+}$ implicitly regularizes the weights of the correctly classified samples with a weighted $\ell_1$ norm term (see supplement). To further enhance sparsity, we may introduce an explicit $\ell_1$ norm regularization term into the projection step with a regularization factor denoted by $\alpha_t\eta_t$. The solution of the projection step is then normalized to get a feasible point on the probability simplex. This algorithm is listed in Algorithm \ref{sparseboost}. $\alpha_t\eta_t$ is the regularization factor of the explicit $\ell_1$ norm at round $t$. Note that the dominant regularization factor is $\eta_t d_t^i$ which only pushes the weights of the correctly classified samples to zero .i.e., when $d_t^i\!<\!0$. This can become evident by substituting the update step in the projection step for $\mathbf{z}_{t+1}$.
  
  For simplicity we consider two cases: when $\alpha_t\!=\text{min}(1,\frac{1}{2}\gamma_t||y_t||_1)\!$ and when $\alpha_t\!=\!0$. The following theorem bounds the training error.
 
\textbf{Theorem 4.} \textit{Suppose that SparseBoost generates weak hypotheses $h_1,\dots,h_T$ whose edges are $\gamma_1,\dots,\gamma_T$. Then the error $\epsilon$ of the combined hypothesis $f$ on the training set is bounded as follows:}
\vspace{-4mm}
\begin{align}
  \label{eq_18}
\epsilon \le \frac{1}{1+c\sum_{t=1}^T  \gamma_t^2||\mathbf{y}_t||_1^2 } 
\vspace{2mm}
   \end{align}
Note that this bound holds for any choice of $\alpha\in\big[0,\text{min}(1,\gamma_t||y_t||_1)\big)$. Particularly, in our two cases constant $c$ is 1 for $\alpha_t\!=\!0$, and $\frac{1}{4}$ when $\alpha_t\!=\text{min}(1,\frac{1}{2}\gamma_t||y_t||_1)\!$.

For $\alpha_t\!=\!0$,  the $\ell_1$ norm of the weights $||y_t||_1$ can be bounded away from zero by $\frac{1}{N}$ (see supplement). Thus, the error $\epsilon$ tends to zero by $O(\frac{N^2}{\gamma^2T})$. That is, in this case Sparseboost is a provable boosting algorithm. However, for $\alpha_t\!\neq\!0$, the $\ell_1$ norm $||y_t||_1$ may rapidly go to zero which consequently results in a non-vanishing upper bound (as $T$ increases) for the training error in \eref{eq_18}. In this case, it may not be possible to conclude that the algorithm is in fact a boosting algorithm\footnote{Nevertheless, for some choices of $\alpha_t\!\neq\!0$ such as  $\alpha_t \propto \frac{1}{t^2}$, the boosting property of the algorithm is still provable.}. It is noteworthy that SparseBoost can be seen as a variant of the COMID algorithm in \cite{duch:10}.
\begin{algorithm}[H]
\label{sparseboost}
%\SetKw{Initialize}{Initialize}
\SetKwInput{Input}{Input}
\SetKwInOut{Output}{Output}
  \caption{SparseBoost}
  Let $\mathcal{R}_{+}\!=\!\{\mathbf{y}\enspace|\enspace0\!\le y_i\}$; Set $\mathbf{y}_1=[\frac{1}{N},\dots,\frac{1}{N}]^{\top}$\; 
  At $t=1,\dots,T$, train $h_t$, set $(\eta_t\!=\!\frac{\gamma_t||y_t||_1}{N}, \alpha_t\!=\!0)$ or $(\eta_t\!=\!\frac{\gamma_t||y_t||_1}{2N}, \alpha_t\!=\!\frac{1}{2}\gamma_t||y_t||_1)$, and update
  \begin{align*}
  & \mathbf{z}_{t+1} = \mathbf{y}_{t}+\eta_t\mathbf{d}_{t} \\
  & \mathbf{y}_{t+1}=\argmin_{\mathbf{y}\in \mathcal{R}_{+}} \frac{1}{2}||\mathbf{y}- \mathbf{z}_{t+1}||^2+\alpha_t\eta_t ||\mathbf{y}||_1\rightarrow y_{t+1}^i=\text{max}(0,y_t^i+\eta_td_t^i-\alpha_t\eta_t)\\
  & \mathbf{w}_{t+1}=\frac{\mathbf{y}_{t+1}}{\sum_{i=1}^N y_i}
   \end{align*}
 Output the final hypothesis $f(\mathbf{x}) \!=\text{sign}\bigg(\sum_{t=1}^T \eta_t h_{t}(\mathbf{x})\bigg)$.
\end{algorithm}
\vspace{-1mm}
\subsection{Lazy Update Boosting}
In this section, we present the proof for the lazy update version of MABoost (LAMABoost) in Theorem 1. The proof technique is novel and can be used to generalize several known online learning algorithms such as OMDA in \cite{rakhlin:13} and Meta algorithm in \cite{chiang:12}. Moreover, we show that MadaBoost \cite{domingo:00} can be presented in the LAMABoost setting. This gives a simple proof for MadaBoost without making the assumption that the edge sequence is monotonically decreasing (as in \cite{domingo:00}). 

\textit{Proof}: Assume $\mathbf{w}^*=[w_1^*,\dots\ ,w_N^*]^{\top}$ is a distribution vector where $w^*_i=\frac{1}{N\epsilon}$ if $f(\mathbf{x}_i)\neq a_i$, and 0 otherwise. Then,
\vspace{-1mm}
\begin{align}
\label{eq_19}
(\mathbf{w}^*\!-\mathbf{w}_{t}&)^{\top}\eta_t\mathbf{d}_{t}\!=(\mathbf{w}_{t+1}-\mathbf{w}_{t})^{\top}\big(\nabla\mathcal{R}(\mathbf{z}_{t+1})-\nabla\mathcal{R}(\mathbf{z}_{t})\big)\nonumber\\ &+(\mathbf{z}_{t+1}-\mathbf{w}_{t+1})^{\top}\big(\nabla\mathcal{R}(\mathbf{z}_{t+1})-\nabla\mathcal{R}(\mathbf{z}_{t})\big)
 +(\mathbf{w}^*-\mathbf{z}_{t+1})^{\top}\big(\nabla\mathcal{R}(\mathbf{z}_{t+1})-\nabla\mathcal{R}(\mathbf{z}_{t})\big)\nonumber\\
&\le  \frac{1}{2}||\mathbf{w}_{t+1}-\mathbf{w}_t||^2+\frac{1}{2}\eta_t^2||\mathbf{d}_t||^2_*+B_{\mathcal{R}}(\mathbf{w}_{t+1},\mathbf{z}_{t+1})-B_{\mathcal{R}}(\mathbf{w}_{t+1},\mathbf{z}_{t})+B_{\mathcal{R}}(\mathbf{z}_{t+1},\mathbf{z}_{t})\nonumber\\
&-B_{\mathcal{R}}(\mathbf{w}^*,\mathbf{z}_{t+1})+B_{\mathcal{R}}(\mathbf{w}^*,\mathbf{z}_{t})-B_{\mathcal{R}}(\mathbf{z}_{t+1},\mathbf{z}_{t})\nonumber\\
&\le \frac{1}{2}||\mathbf{w}_{t+1}-\mathbf{w}_t||^2+\frac{1}{2}\eta_t^2||\mathbf{d}_t||^2_*-B_{\mathcal{R}}(\mathbf{w}_{t+1},\mathbf{w}_{t})\nonumber\\
&+B_{\mathcal{R}}(\mathbf{w}_{t+1},\mathbf{z}_{t+1})-B_{\mathcal{R}}(\mathbf{w}_{t},\mathbf{z}_{t})-B_{\mathcal{R}}(\mathbf{w}^*,\mathbf{z}_{t+1})+B_{\mathcal{R}}(\mathbf{w}^*,\mathbf{z}_{t})
\end{align}
where the first inequality follows applying Lemma 3 to the first term and Lemma 2 to the rest of the terms and the second inequality is the result of applying the exact version of Lemma 1 to $B_{\mathcal{R}}(\mathbf{w}_{t+1},\mathbf{z}_{t})$. Moreover, since $B_{\mathcal{R}}(\mathbf{w}_{t+1},\mathbf{w}_{t})-\!\frac{1}{2}||\mathbf{w}_{t+1}-\mathbf{w}_t||^2\ge0$, they can be ignored in \eref{eq_19}. Summing up the inequality \eref{eq_19} from $t=1$ to $T$, yields
\begin{align} 
\label{eq_20}
  -B_{\mathcal{R}}(\mathbf{w}^*,\mathbf{z}_{1})\le L\sum_{t=1}^T \frac{1}{2}\eta_t^2  - \sum_{t=1}^T \eta_t\gamma_t 
\end{align}
where we used the facts that $\mathbf{w}^{*\top}\sum_{t=1}^T \eta_t \mathbf{d}_t\!\ge0$ and $\sum_{t=1}^T -\mathbf{w}_t^{\top}\eta_t\mathbf{d}_t\!=\sum_{t=1}^T \eta_t\gamma_t$. The above inequality is exactly the same as \eref{eq_15}, and replacing $-B_{\mathcal{R}}$ with $\frac{\epsilon-1}{N\epsilon}$ or $\log(\epsilon)$ yields the same error bounds in Theorem 1. Note that, since the exact version of Lemma 1 is required to obtain \eref{eq_19}, this proof does not reveal whether LAMABoost can be generalized to employ the double-projection strategy. In some particular cases, however, we may show that a double-projection variant of LAMABoost is still a provable boosting algorithm. 

In the following, we briefly show that MadaBoost can be seen as a double-projection LAMABoost. 
%
 % \begin{algorithm}[H]
%\SetKw{Initialize}{Initialize}
%\SetKwInput{Input}{Input}
%\SetKwInOut{Output}{Output}
%  \caption{Variant of MadaBoost}
%   Set $\mathbf{w}_1=\mathbf{z}_1=\mathbf{1}$ and $\epsilon$ to the desired error. At $t=1,\dots,T$, train $h_t$ with $\mathbf{w}_t/||\mathbf{w}_t||_1$, set $\eta_t=\gamma_t$ and update: 
%   \quad $z^i_{t+1} = z^i_t e^{\eta_t d_{t}^i}, \quad w_{t+1}^i=\min(1,z^i_{t+1})$
   
% Output the final hypothesis $f(\mathbf{x}) \!=\text{sign}\bigg(\sum_{t=1}^T \eta_t h_{t}(\mathbf{x})\bigg)$.
%\end{algorithm}

 \begin{algorithm}[H]
%\SetKw{Initialize}{Initialize}
\SetKwInput{Input}{Input}
\SetKwInOut{Output}{Output}
  \caption{Variant of MadaBoost}
  Let $\mathcal{R}(\mathbf{w})$ be the negative entropy and $\mathcal{K}$ a unit hypercube; Set $\mathbf{z}_1=[1,\dots,1]^\top$\; 
  At $t=1,\dots,T$, train $h_t$ with $\mathbf{w}_t$, set $f_t(\mathbf{x})\!=\text{sign}\bigg(\sum_{t'=1}^t \eta_{t'} h_{t'}(\mathbf{x})\bigg)$ and calculate $\epsilon_t=\frac{\mathlarger{\sum_{i=1}^N \frac{1}{2}|f_t(\mathbf{x}_i)- a_i|}}{N}$, set $\eta_t=\epsilon_t\gamma_t$ and update
  \begin{align*}
  &\nabla\mathcal{R}(\mathbf{z}_{t+1}) = \nabla\mathcal{R}(\mathbf{z}_{t})+\eta_t\mathbf{d}_{t} &&\rightarrow z^i_{t+1} = z^i_t e^{\eta_t d_{t}^i} \\
  & \mathbf{y}_{t+1}=\argmin_{\mathbf{y}\in \mathcal{K}} B_{\mathcal{R}}(\mathbf{y},\mathbf{z}_{t+1})  &&\rightarrow y_{t+1}^i=\min(1,z^i_{t+1}) \\
  &\mathbf{w}_{t+1}=\argmin_{\mathbf{w}\in \mathcal{S}} B_{\mathcal{R}}(\mathbf{w},\mathbf{y}_{t+1}) &&\rightarrow w_{t+1}^i=\frac{y_{t+1}^i }{||\mathbf{y}_{t+1}||_1}
   \end{align*}
 Output the final hypothesis $f(\mathbf{x}) \!=\text{sign}\bigg(\sum_{t=1}^T \eta_t h_{t}(\mathbf{x})\bigg)$.
\end{algorithm}
Algorithm 3 is essentially MadaBoost, only with a different choice of $\eta_t$. It is well-known that the entropy projection onto the probability simplex results in the normalization and thus, the second projection of Algorithm 3. The entropy projection onto the unit hypercube, however, maybe less known and thus, its proof is given in the Supplement.

\textbf{Theorem 5.} \textit{Algorithm 3 yields a $(1\!-\epsilon)$-accurate hypothesis after at most $T\!=O(\frac{1}{\mathlarger{\epsilon^2\gamma^2}})$}.

This is an important result since it shows that MadaBoost seems, at least in theory, to be slower than what we hoped, namely $O(\frac{1}{\mathlarger{\epsilon\gamma^2}})$.
\vspace{-2mm}
\section{Conclusion and Discussion}
In this work, we provided a boosting framework that can produce provable boosting algorithms. This framework is mainly suitable for designing boosting algorithms with distribution constraints. A sparse boosting algorithm that samples only a fraction of examples at each round was derived from this framework. However, since our proposed algorithm cannot control the exact number of zeros in the weight vector, a natural extension to this algorithm is to develop a boosting algorithm that receives the sparsity level as an input. However, this immediately raises the question: what is the maximum number of examples that can be removed at each round from the dataset, while still achieving a $(1\!-\epsilon)$-accurate hypothesis?

The boosting framework derived in this work is essentially the dual of the online mirror descent algorithm. This framework can be generalized in different ways. Here, we showed that replacing the Bregman projection step with the double-projection strategy, or as we call it approximate Bregman projection, still results in a boosting algorithm in the active version of MABoost, though this may not hold for the lazy version. In some special cases (MadaBoost for instance), however, it can be shown that this double-projection strategy works for the lazy version as well. Our conjecture is that under some conditions on the first convex set, the lazy version can also be generalized to work with the approximate projection operator. Finally, we provided a new error bound for the MadaBoost algorithm that does not depend on any assumption. Unlike the common conjecture, the convergence rate of MadaBoost (at least with our choice of $\eta$) is of $O(1/\epsilon^2)$. 
\vspace{-2mm}
\subsubsection*{Acknowledgments}
This work was partially supported by SNSF. We would like to thank Professor Rocco Servedio for an inspiring email conversation and our colleague Hui Liang for his helpful comments.
\newpage
\small{

\bibliographystyle{plain}
\bibliography{/home/tofighn/latex/bib}

\begin{thebibliography}{10}

\bibitem{schapire:90}
R.~E. Schapire.
\newblock The strength of weak learnability.
\newblock {\em Mach. Learn.}, 1990.

\bibitem{freund:97}
Y.~Freund and R.~E. Schapire.
\newblock A decision-theoretic generalization of on-line learning and an
  application to boosting.
\newblock {\em Journal of Computer and System Sciences}, 1997.

\bibitem{breiman:99}
L.~Breiman.
\newblock Prediction games and arcing algorithms.
\newblock {\em Neural Comput.}, 1999.

\bibitem{mason:99}
L.~Mason, J.~Baxter, P.~Bartlett, and M.~Frean.
\newblock Boosting algorithms as gradient descent.
\newblock In {\em {NIPS}}, 1999.

\bibitem{friedman:00}
J.~Friedman, T.~Hastie, and R.~Tibshirani.
\newblock Additive logistic regression: a statistical view of boosting.
\newblock {\em Annals of Statistics}, 1998.

\bibitem{servedio:03}
R.~A. Servedio.
\newblock Smooth boosting and learning with malicious noise.
\newblock {\em J. Mach. Learn. Res.}, 2003.

\bibitem{gavinsky:03}
D.~Gavinsky.
\newblock Optimally-smooth adaptive boosting and application to agnostic
  learning.
\newblock {\em J. Mach. Learn. Res.}, 2003.

\bibitem{bradley:08}
J.~K. Bradley and R.~E. Schapire.
\newblock Filterboost: Regression and classification on large datasets.
\newblock In {\em NIPS}. 2008.

\bibitem{hatano:06}
K.~Hatano.
\newblock Smooth boosting using an information-based criterion.
\newblock In {\em Algorithmic Learning Theory}. 2006.

\bibitem{domingo:00}
C.~Domingo and O.~Watanabe.
\newblock Madaboost: A modification of {A}da{B}oost.
\newblock In {\em COLT}, 2000.

\bibitem{freund:95}
Y.~Freund.
\newblock Boosting a weak learning algorithm by majority.
\newblock 1995.

\bibitem{bshouty:02}
N.~H. Bshouty, D.~Gavinsky, and M.~Long.
\newblock On boosting with polynomially bounded distributions.
\newblock {\em Journal of Machine Learning Research}, 2002.

\bibitem{warmuth:06}
M.~K. Warmuth, J.~Liao, and G.~R\"{a}tsch.
\newblock Totally corrective boosting algorithms that maximize the margin.
\newblock In {\em ICML}, 2006.

\bibitem{collins:02}
M.~Collins, R.~E. Schapire, and Y.~Singer.
\newblock Logistic regression, adaboost and bregman distances.
\newblock {\em Mach. Learn.}, 2002.

\bibitem{shwartz:08}
S.~Shalev-Shwartz and Y.~Singer.
\newblock On the equivalence of weak learnability and linear separability: new
  relaxations and efficient boosting algorithms.
\newblock In {\em COLT}, 2008.

\bibitem{zinkevich:03}
M.~Zinkevich.
\newblock Online convex programming and generalized infinitesimal gradient
  ascent.
\newblock In {\em {ICML}}, 2003.

\bibitem{hazan:09}
E.~Hazan.
\newblock A survey: The convex optimization approach to regret minimization.
\newblock Working draft, 2009.

\bibitem{duch:10}
J.~C. Duchi, S.~Shalev-shwartz, Y.~Singer, and A.~Tewari.
\newblock Composite objective mirror descent.
\newblock In {\em {COLT}}, 2010.

\bibitem{freund96}
Y.~Freund and R.~E. Schapire.
\newblock Game theory, on-line prediction and boosting.
\newblock In {\em COLT}, 1996.

\bibitem{breiman:97}
L.~Breiman.
\newblock Pasting bites together for prediction in large data sets and on-line.
\newblock Technical report, Dept. Statistics, Univ. California, Berkeley, 1997.

\bibitem{kearns:92}
M.~J. Kearns, R.~E. Schapire, and L.~M. Sellie.
\newblock Toward efficient agnostic learning.
\newblock In {\em COLT}, 1992.

\bibitem{kalai:09}
A.~Kalai and V.~Kanade.
\newblock Potential-based agnostic boosting.
\newblock In {\em NIPS}. 2009.

\bibitem{ben:01}
S.~Ben-David, P.~Long, and Y.~Mansour.
\newblock Agnostic boosting.
\newblock In {\em Computational Learning Theory}. Springer, 2001.

\bibitem{bianchi:06}
N.~Cesa-Bianchi and G.~Lugosi.
\newblock {\em Prediction, Learning, and Games}.
\newblock Cambridge University Press, 2006.

\bibitem{dai:07}
W.~Dai, Q.~Yang, G.~Xue, and Y.~Yong.
\newblock Boosting for transfer learning.
\newblock In {\em {ICML}}, 2007.

\bibitem{wang:13}
W.~Wang and M.~A. Carreira-Perpi{\~n}{\'a}n.
\newblock Projection onto the probability simplex: An efficient algorithm with
  a simple proof, and an application.
\newblock arXiv:1309.1541, 2013.

\bibitem{rakhlin:13}
A.~Rakhlin and K.~Sridharan.
\newblock Online learning with predictable sequences.
\newblock In {\em COLT}, 2013.

\bibitem{chiang:12}
C.~Chiang, T.~Yang, C.~Lee, M.~Mahdavi, C.~Lu, R.~Jin, and S.~Zhu.
\newblock Online optimization with gradual variations.
\newblock In {\em COLT}, 2012.

\bibitem{freund:96}
Y.~Freund and R.~E. Schapire.
\newblock {Experiments with a New Boosting Algorithm}.
\newblock In {\em ICML}, 1996.

\bibitem{rw05}
G.~R\"{a}tsch and M.~Warmuth.
\newblock Efficient margin maximization with boosting.
\newblock {\em Journal of Machine Learning Research}, 2005.

\bibitem{shalev:12}
Shai Shalev-Shwartz.
\newblock Online learning and online convex optimization.
\newblock {\em Found. Trends Machine Learning}, 2012.

\end{thebibliography}
}

\newpage

\small{

\bibliographystyle{plain}

}
\newpage
\section* {Supplement}
 Before proceeding with the proofs, some definitions and facts need to be reminded.
 
 \subsubsection*{Definition: Margin} \textit{Given a final hypothesis $f(\mathbf{x}) \!=\! \sum_{t=1}^T \eta_t h_{t}(\mathbf{x})$, the margin of a sample $(\mathbf{x}_j,a_j)$ is defined as $m(\mathbf{x}_j)\!=\!a_jf(\mathbf{x}_j)/\sum_{t=1}^T \eta_t $. Moreover, the margin of a set of examples denoted by $m_\mathcal{D}$ is the minimum of margins over the examples, i.e., $m_\mathcal{D} \!=\! \min_{\mathbf{x}} m(\mathbf{x}_j)$}. 

\subsubsection*{Fact: Duality between max-margin and min-edge} \textit{The minimum edge $\gamma_{\text{min}}$ that can be achieved over all possible distributions of the training set is equal to the maximum margin ($m^*=\max_{\eta} m_\mathcal{D}$) of any linear combination of hypotheses from the hypotheses space}.

This fact is discussed in details in \cite{freund:96} and \cite{rw05}. It is the direct result of von Neumann's minmax theorem and simply means that the maximum achievable margin is $\gamma_{\text{min}}$. 

\subsection*{\textbf{Proof of Theorem 2}} The proof for the maximum margin property of MABoost, is almost the same as the proof of Theorem 1.

Let's assume the $i^\text{th}$ sample has the worst margin, i.e., $m_\mathcal{D}= m(\mathbf{x}_i)$. Let all entries of the error vector $\mathbf{w}^*$ to be zero except its $i^\text{th}$ entry which is set to be 1. Following the same approach as in Theorem 1, (see equation \eref{eq_13}), we get
\begin{align} 
\label{apT2_1}
 \sum_{t=1}^{T} \mathbf{w}^{*\top}\eta_t\mathbf{d}_{t}\!-\mathbf{w}_{t}^{\top}\eta_t\mathbf{d}_{t}\le \sum_{t=1}^T \frac{1}{2}\eta_t^2||\mathbf{d}_{t}||^2_*  +   B_{\mathcal{R}}(\mathbf{w}^*,\mathbf{w}_{1}) -  B_{\mathcal{R}}(\mathbf{w}^*,\mathbf{w}_{T+1}) 
\end{align}
With our choice of $\mathbf{w}^*$ it is easy to verify that the first term on the left side of the inequality is $m_\mathcal{D}\sum_{t=1}^{T} \eta_t \!=\! -\sum_{t=1}^{T} \mathbf{w}^{*\top}\eta_t\mathbf{d}_{t}$.
By setting $C=B_{\mathcal{R}}(\mathbf{w}^*,\mathbf{w}_{1})$, ignoring the last term $B_{\mathcal{R}}(\mathbf{w}^*,\mathbf{w}_{T+1})$, replacing $||\mathbf{d}_{t}||^2_*$ with its upper bound $L$ and using the identity  $\sum_{t=1}^T \mathbf{w}_t^{\top}\eta_t\mathbf{d}_t\!=-\sum_{t=1}^T \eta_t\gamma_t$ the above inequality is simplified to
\begin{align} 
\label{apT2_2}
  -m_\mathcal{D}\sum_{t=1}^{T} \eta_t \le L\sum_{t=1}^T \frac{1}{2}\eta_t^2  -\sum_{t=1}^T \eta_t\gamma_t +   C 
\end{align}
Replacing $\eta_t$ with the value suggested in Theorem 2, i.e., $\eta_t=\mathlarger{\frac{\gamma_t}{L\sqrt{t}}}$ and dividing both sides by $\sum_{t=1}^{T} \eta_t$, gives
\begin{align} 
\label{apT2_2}
    \frac{\sum_{t=1}^T(\frac{1}{\sqrt{t}}-\frac{1}{t})\gamma_t^2}{\sum_{t=1}^T \frac{1}{\sqrt{t}}\gamma_t}  - \frac{LC}{{\sum_{t=1}^T\frac{1}{\sqrt{t}}\gamma_t}} \le m_\mathcal{D}
\end{align}
The first term is minimized when $\gamma_t\!=\!\gamma_{\text{min}}$ . Similarly to the first term, the second term is maximized when $\gamma_t$ is replaced by its minimum value. This gives the following lower bound for $m_\mathcal{D}$:
\begin{align} 
\label{apT2_3}
 \gamma_{\text{min}}  \frac{ \sum_{t=1}^T \frac{1}{\sqrt{t}}-\frac{1}{t}}{\sum_{t=1}^T \frac{1}{\sqrt{t}}}  - \frac{LC}{\gamma_{\text{min} }\sum_{t=1}^T \frac{1}{\sqrt{t}} } \le m_\mathcal{D} 
\end{align}
Considering the facts that $\int_1^{T+1}\frac{dx}{\sqrt{x}}  \le \sum_{t=1}^T \frac{1}{\sqrt{t}}$ and $1+\int_1^{T} \frac{dx}{x}  \ge \sum_{t=1}^T \frac{1}{t}$, we get
\begin{align} 
\label{apT2_3}
 \gamma_{\text{min}} - \frac{1+\log T}{2\sqrt{T+1}-2} \gamma_{\text{min}} - \frac{LC}{\gamma_{\text{min} }(\sqrt{T+1}-1) } \le m_\mathcal{D}
\end{align}
Now by taking $\nu=\frac{1+\log T}{2\sqrt{T+1}-2} \gamma_{\text{min}} + \frac{LC}{\gamma_{\text{min} }(\sqrt{T+1}-1) } $, we have $\gamma_{\text{min}} -\nu \le \gamma_{\text{min}}$. It is clear from \eref{apT2_3} that $\nu$ approaches zero as $T$ tends to infinity with a convergence rate proportional to $\frac{\log T}{\sqrt{T}}$. It is noteworthy that this convergence rate is slightly worse than that of TotalBoost which is $O(\frac{1}{\sqrt{T}})$.

\subsection*{\textbf{Proof of Lemma 4}} Remember that $\hat{\Pi}_{\mathcal{S}}(\mathbf{y})\!=\Pi_{\mathcal{S}}\Big(\Pi_{\mathcal{K}}(\mathbf{y})\Big)$. Our goal is to show that $B_{\mathcal{R}}(\mathbf{x},\mathbf{y})\ge B_{\mathcal{R}}\big(\mathbf{x},\hat{\Pi}_{\mathcal{S}}(\mathbf{y})\big)$.

To this end, we only need to repeatedly apply Lemma 1, as follows
 \begin{align}
 &B_{\mathcal{R}}(\mathbf{x},\mathbf{y})\ge B_{\mathcal{R}}\big(\mathbf{x},\Pi_{\mathcal{K}}(\mathbf{y})\big)\\
 &B_{\mathcal{R}}\big(\mathbf{x},\Pi_{\mathcal{K}}(\mathbf{y})\big)\ge  B_{\mathcal{R}}\big(\mathbf{x},\hat{\Pi}_{\mathcal{S}}(\mathbf{y})\big)
 \end{align}
which completes the proof.

\subsection*{\textbf{Proof of combining datasets boosting algorithm}} We have to show that when the convex set is defined as
\begin{align}
\mathcal{S}_c\!=\{\mathbf{w}|\sum_{i=1}^N w_i\!=1 , 0\!\le w_i\enskip \forall i\in\mathcal{A} \enskip \wedge \enskip 0\!\le w_i\!\le \frac{k}{N}\enskip \forall i\in\mathcal{B}  \}
\end{align}
the error of the final hypothesis on $\mathcal{A}$, i.e., $\epsilon_{\mathcal{A}}$, converges to zero while the error on $\mathcal{B}$ is guaranteed to be $\epsilon_{\mathcal{B}}\le \frac{1}{k}$. 

First, we show the convergence of $\epsilon_{\mathcal{A}}$ to zero. This is easily obtained by setting $\mathbf{w}^*$ to be an error vector with zero weights over the training samples from $\mathcal{B}$ and $\frac{1}
{\mathlarger{\epsilon}_{\mathcal{A}} N_{\mathcal{A}}}$ weights over the training set $\mathcal{A}$. One can verify that $w^*\in \mathcal{S}_c$, thus the proof of Theorem 1 holds and subsequently, the error bounds in \eref{eq_10} stating that $\epsilon_{\mathcal{A}}\rightarrow 0 $ as the number of iterations increases. 

To show the second part of the theorem that is $\epsilon_{\mathcal{B}}\le \frac{1}{k}$, vector $\mathbf{w}^*$ is selected to be an error vector with zero weights over the training samples from $\mathcal{A}$ and $\frac{1}
{\mathlarger{\epsilon}_{\mathcal{B}} N_{\mathcal{B}}}$ weights over the training set $\mathcal{B}$. Note that, as long as $\epsilon_{\mathcal{B}}$ is greater than $\frac{1}{k}$, this $w^*\in \mathcal{S}_c$. Thus, for all $\frac{1}{k}\le \epsilon_{\mathcal{B}}$ the proof of Theorem 1 holds and as the bounds in \eref{eq_10} show, the error decreases as the number of iterations increases. In particular in a finite number of rounds, the classification error on $\mathcal{B}$ reduces to $\frac{1}{k}$ which completes the proof.

\subsection*{\textbf{Proof of Theorem 4}}
We use proof techniques similar to those given in \cite{duch:10}, with a slight change to take the normalization step into account.

By replacing $\mathbf{z}_{t+1}$ in the projection step from the update step, the projection step can be rewritten as
\begin{align}
\label{t4_1}
\mathbf{y}_{t+1}=\argmin_{\mathbf{y}\in \mathcal{R}_+} \frac{1}{2}||\mathbf{y}-\mathbf{y}_t|| -\eta_t\mathbf{y}^\top \mathbf{d}_t + \alpha_t\eta_t||\mathbf{y}||_1
\end{align}
This optimization problem can be highly simplified by noting that the variables are not coupled. Thus, each coordinate can be independently optimized. In other words, it can be decoupled into $N$ independent 1-dimensional optimization problems.
\begin{align}
\label{t4_2}
y_{t+1}^i=\argmin_{0\le y_i} \frac{1}{2}||y_i-y_t^i|| -\eta_t y_id_t^i + \alpha_t\eta_ty_i
\end{align}
The solution of \eref{t4_2} can be written as
\begin{align}
\label{t4_3}
y_{t+1}^i=\text{max}(0,y_t^i+\eta_td_t^i - \alpha_t\eta_t)
\end{align}
This simple solution gives a very efficient and simple implementation for SparseBoost. From \eref{t4_2} it is clear that for $d_t^i<0$ (i.e., when $i{^\text{th}}$ sample is classified correctly), $-\eta_t yd_t^i$ acts as the $\ell_1$ norm regularization and pushes $y_{t+1}^i$ towards zero while $\alpha_t\eta_t$ enhance sparsity by pushing all weights to zero.

Let $\mathbf{w}^*$ to be the same error vector as defined in Theorem 1. We start this proof by again deriving the progress bounds on each step of the algorithm. The optimality of $\mathbf{y}_{t+1}$ for \eref{t4_1} implies that
\begin{align}
\label{t4_4}
(\mathbf{w}^*-\mathbf{y}_{t+1})^\top(-\eta_t\mathbf{d}_t+\alpha_t\eta_t r'(\mathbf{y}) + \mathbf{y}_{t+1} - \mathbf{y}_t)\ge0 
\end{align}
where $r'(\mathbf{y})$ is a sub-gradient vector of the $\ell_1$ norm function $r(\mathbf{y})=\sum_{i=1}^N y_i$. Moreover, due to the convexity of $r(\mathbf{y})$, we have
\begin{align}
\label{t4_5}
\alpha_t\eta_t r(\mathbf{y}_{t+1})^\top(\mathbf{w}^*-\mathbf{y}_{t+1})\le \alpha_t\eta_t \big(r(\mathbf{w}^*)-r(\mathbf{y}_{t+1})\big)
\end{align}
We thus have
\begin{align}
\label{t4_6}
&(\mathbf{w}^*\!-\mathbf{y}_{t})^{\top}\eta_t\mathbf{d}_{t} + \alpha_t\eta_t\big(r(\mathbf{y}_{t+1}) -r(\mathbf{w}^*)\big)\le
(\mathbf{w}^*\!-\mathbf{y}_{t})^{\top}\eta_t\mathbf{d}_{t} + \alpha_t\eta_t(\mathbf{y}_{t+1}-\mathbf{w}^*)^\top r'(\mathbf{y}_{t+1})\nonumber \\
&=(\mathbf{w}^*\!-\mathbf{y}_{t+1})^{\top}\eta_t\mathbf{d}_{t} + \alpha_t\eta_t(\mathbf{y}_{t+1}-\mathbf{w}^*)^\top r'(\mathbf{y}_{t+1})
+(\mathbf{y}_{t+1}-\mathbf{y}_{t})^\top \eta_t\mathbf{d}_{t} \nonumber\\
&=(\mathbf{w}^*\!-\mathbf{y}_{t+1})^{\top}(\eta_t\mathbf{d}_{t} -\alpha_t\eta_t r'(\mathbf{y}_{t+1})-\mathbf{y}_{t+1} + \mathbf{y}_t)\nonumber\\
&+(\mathbf{w}^*\!-\mathbf{y}_{t+1})^{\top}(\mathbf{y}_{t+1} - \mathbf{y}_t) +(\mathbf{y}_{t+1}-\mathbf{y}_{t})^\top \eta_t\mathbf{d}_{t}
\end{align}
where the first inequality follows \eref{t4_5}. Now, from the optimality condition in \eref{t4_4}, the first term in the last equation is non-positive and thus, can be ignored.
\begin{align}
\label{t4_7}
&(\mathbf{w}^*\!-\mathbf{y}_{t})^{\top}\eta_t\mathbf{d}_{t} + \alpha_t\eta_t\big(r(\mathbf{y}_{t+1}) -r(\mathbf{w}^*)\big)\le
(\mathbf{w}^*\!-\mathbf{y}_{t+1})^{\top}(\mathbf{y}_{t+1} - \mathbf{y}_t) +(\mathbf{y}_{t+1}-\mathbf{y}_{t})^\top \eta_t\mathbf{d}_{t} \nonumber\\
&=\frac{1}{2}||\mathbf{w}^*\!-\mathbf{y}_{t}||_2^2-\frac{1}{2}||\mathbf{y}_{t+1}\!-\mathbf{y}_{t}||_2^2-\frac{1}{2}||\mathbf{w}^*\!-\mathbf{y}_{t+1}||_2^2 + (\mathbf{y}_{t+1}-\mathbf{y}_{t})^\top \eta_t\mathbf{d}_{t} \nonumber\\
&\le \frac{1}{2}||\mathbf{w}^*\!-\mathbf{y}_{t}||_2^2-\frac{1}{2}||\mathbf{y}_{t+1}\!-\mathbf{y}_{t}||_2^2-\frac{1}{2}||\mathbf{w}^*\!-\mathbf{y}_{t+1}||_2^2 +\frac{1}{2}||\mathbf{y}_{t+1}\!-\mathbf{y}_{t}||_2^2+ \frac{1}{2} \eta_t^2||\mathbf{d}_t||_*^2
\end{align}
where the first equation follows from Lemma 2 (or direct algebraic expansion in this case) and the second inequality from Lemma 3. 

By summing the left and right sides of the inequality from $1$ to $T$, replacing $||\mathbf{d}_t||_*^2$ with its upperbound $N$ and substituting 1 for $r(\mathbf{w}^*)$, we get
\begin{align}
\label{t4_8}
 \sum_{t=1}^{T} \mathbf{w}^{*\top}\eta_t\mathbf{d}_{t}\le \sum_{t=1}^{T} \mathbf{y}_{t}^{\top}\eta_t\mathbf{d}_{t}+ \sum_{t=1}^T \frac{N}{2}\eta_t^2 + \frac{1}{2}||\mathbf{w}^*\!-\mathbf{y}_{1}||_2^2
                                                               +\sum_{t=1}^T\alpha_t\eta_t\big(1-r(\mathbf{y}_{t+1})\big)
\end{align}
Now, replacing $r(\mathbf{y}_{t+1})$ with its lower bound, i.e, 0 and using the fact that $\sum_{t=1}^{T} \mathbf{w}^{*\top}\eta_t\mathbf{d}_{t}\ge0$ (as shown in \eref{eq_14}) and $\sum_{t=1}^T \mathbf{y}_t^{\top}\eta_t\mathbf{d}_t\!=-\sum_{t=1}^T \eta_t\gamma_t ||\mathbf{y}_t||_1$, yields
\begin{align}
\label{t4_9}
 0\le -\sum_{t=1}^T \eta_t\gamma_t||\mathbf{y}_t||_1+ \sum_{t=1}^T \frac{N}{2}\eta_t^2+ \frac{1}{2}||\mathbf{w}^*\!-\mathbf{y}_{1}||_2^2
                                                               +\sum_{t=1}^T\alpha_t\eta_t
\end{align}
Taking derivative w.r.t $\eta_t$ and setting it to zero, gives the optimal $\eta_t$ as follows
\begin{align}
\label{t4_10}
 \eta_t = \frac{\gamma_t||\mathbf{y}_t||_1-\alpha_t}{N}
\end{align}
This equation implies that $\alpha_t$ should be smaller than $\gamma_t||\mathbf{y}_t||_1$ or otherwise $\eta_t$ becomes smaller than zero. Setting $\alpha_t=(1-k)\gamma_t||\mathbf{y}_t||_1$
where $k$ is a constant smaller than or equal to 1, results in $\eta_t=\frac{k}{N}\gamma_t||\mathbf{y}_t||_1$. Replacing this value for $\eta_t$ in \eref{t4_9} and noting that $\frac{1}{2}||\mathbf{w}^*\!-\mathbf{y}_{1}||_2^2=\frac{1-\epsilon}{2N\epsilon}$ gives the following bound on the training error
\begin{align}
\label{t4_11}
\epsilon \le \frac{1}{1+c\sum_{t=1}^T  \gamma_t^2||\mathbf{y}_t||_1^2 } 
\end{align}
where $c=\frac{1}{k^2}$ is a constant factor depending on the choice of $\alpha_t$. To prove that $\epsilon$ approaches zero as $T$ increases, we still have to provide an evidence that $\sum_{t=1}^T \gamma_t^2||\mathbf{y}_t||_1^2$ is a divergent series. There are different possibilities to approach this problem. Here, we show that in the case of $\alpha_t\!=\!0$, the $\ell_1$ norm of weights $||\mathbf{y}_t||_1$ can be bounded away from zero (i.e., $||\mathbf{y}_t||_1\ge C>0$) and thus, $\sum_{t=1}^T \gamma_t^2||\mathbf{y}_t||_1^2 \ge T\gamma^2_{\text{min}}C^2$. 

To this end, we rewrite $y_t^i$ from \eref{t4_3} as
\begin{align}
\label{t4_12}
y_t^i =\text{max}&(0,y_{t-1}^i+\eta_{t-1}d_{t-1}^i - \alpha_{t-1}\eta_{t-1})\nonumber \\
&\ge  y_{t-1}^i+\eta_{t-1}d_{t-1}^i - \alpha_{t-1}\eta_{t-1} \nonumber\\ 
&\ge \frac{1}{N} +\sum_{t'=1}^{t-1} \eta_{t'}d_{t'}- \sum_{t'=1}^{t-1}\alpha_{t'}\eta_{t'}
\end{align}
where the last inequality is achieved by recursively applying the first inequality to $y_{t-1}^i$. 
At any arbitrary round $t$, either the algorithm has already converged and $\epsilon\!=\!0$ or there is at least one sample that is classified wrongly by the ensemble classifier $H_t(\mathbf{x})=\sum_{l=1}^{t} \eta_l h_{l}(\mathbf{x})$. Now, without loss of generality, assume the $i^{\text{th}}$ sample is wrongly classified at round $t$. That is, $\sum_{t'=1}^{t-1} \eta_{t'}d_{t'}>0$ (look at \eref{eq_14}). Now, for $\alpha_t\!=\!0$, the weight of the wrongly classified sample $i$ is
\begin{align}
\label{t4_13}
y_t^i \ge \frac{1}{N} +\sum_{t'=1}^{t-1} \eta_{t'}d_{t'}\ge\frac{1}{N}
\end{align}
That is, $||\mathbf{y}_t||_1\ge \frac{1}{N}$. This gives a lousy (but sufficient for our purpose) lower bound on $||\mathbf{y}_t||_1$. Replacing $||\mathbf{y}_t||_1$ with its lower bound $\frac{1}{N}$ in \eref{t4_11}, yields
\begin{align}
\label{t4_14}
\epsilon \le \frac{N^2}{1+T  \gamma^2 } 
\end{align}
where $ \gamma$ is the minimum edge over all $\gamma_t$.

\subsection*{\textbf{Proof of Entropy Projection onto Hypercube (Second Update Step in MadaBoost) }}

\textbf{Lemma 5.} \textit{Let $\mathcal{R}(\mathbf{w})\!=\sum_{i=1}^{N} w_i\log{w_i}{-}w_i$. Then the Bregman projection of a positive vector $\mathbf{z}\in\mathbb{R}_+^N$ onto the unit hypercube $\mathcal{K}=[0,1]^N$ is  $y_i=\min(1,z_i),i=1,\dots,N$}.

To show the correctness of the above lemma, i.e., that the solution of the Bregman projection
\begin{align}
\label{ap_l5_1}
 \mathbf{y}=\argmin_{\mathbf{y}\in \mathcal{K}} B_{\mathcal{R}}(\mathbf{y},\mathbf{z})  
\end{align}
is $y_i=\min(1,z_i)$, we only need to show that $\mathbf{y}$ satisfies the optimality condition 
\begin{align}
\label{ap_l5_2}
(\mathbf{v}-\mathbf{y})^\top \nabla B_{\mathcal{R}}(\mathbf{y},\mathbf{z}) \ge0 \quad \forall \mathbf{v}\in \mathcal{K}
\end{align}
Given $\mathcal{R}(\mathbf{w})\!=\sum_{i=1}^{N} w_i\log{w_i}{-}w_i$, the gradient of $B_{\mathcal{R}}$ is
\begin{align}
\label{ap_l5_3}
 \nabla B_{\mathcal{R}}(\mathbf{y},\mathbf{z})=\sum_{i=1}^T \log \frac{y_i}{z_i}
\end{align}
Hence,
\begin{align}
\label{ap_l5_2}
(\mathbf{v}-\mathbf{y})^\top \nabla B_{\mathcal{R}}(\mathbf{y},\mathbf{z})=\sum_{i \in \{i:z_i\ge1\} } (v_i-y_i)\log \frac{y_i}{z_i} + \sum_{i \in \{i:z_i<1\} } (v_i-y_i)\log \frac{y_i}{z_i}
\end{align}
For $z_i\ge1$, $y_i$ is equal to 1. That is, $\log \frac{y_i}{z_i}=\log \frac{1}{z_i}<0$. On the other hand, since $v_i\le1$, $(v_i-y_i)= (v_i-1)\le0$. Thus, the first sum in \eref{ap_l5_2} is always non-negative. The second sum is always zero since $\log \frac{y_i}{z_i}=\log 1=0$. That is, the optimality condition \eref{ap_l5_2} is non-negative for all $\mathbf{v}$ which completes the proof.
 
\subsection*{\textbf{Proof of Theorem 5}} Its proof is essentially the same as the proof of the lazy version of MABoost with a few differences. Before proceeding further, some definitions and facts should be re-emphasized.

First of all, since $\mathcal{R}(\mathbf{w})=\sum_{i=1}^{N}w_i\log w_i-w_i$ is $\frac{1}{N}$-strongly convex (see \cite[p. 136]{shalev:12}) with respect to $\ell_1$ norm (and not 1-strongly as in Theorem 1), the following inequality holds for the Bregman divergence:
\begin{align}
\label{ap_t5_1}
B_{\mathcal{R}}(\mathbf{x},\mathbf{y})\ge \frac{1}{2N}||\mathbf{x}-\mathbf{y}||_1^2
\end{align}
Moreover, the following lemma which bounds $||\mathbf{y}_t||$ is essential for our proof.

\textbf{Lemma 6.} \textit{For all $t$, $||\mathbf{y}_t||_1 \ge N\epsilon_t$ where $\epsilon_t$ is the error of the ensemble hypothesis $H_t(\mathbf{x})=\sum_{l=1}^{t} \eta_l h_{l}(\mathbf{x})$ at round $t$}.

This lemma holds due to the fact that
\begin{align}
\label{ap_t5_2}
y_t^i=\min(1,z_t^i)=\min(1,e^{\sum_{l=1}^{t} \eta_l d_{l}^i})=\min(1,e^{-a_iH_t(\mathbf{x}_i)})
\end{align}
where $H_t(\mathbf{x})=\sum_{l=1}^{t} \eta_l h_{l}(\mathbf{x})$ is the output of the algorithm at round $t$. If $H_t(\mathbf{x}_i)$ makes a mistake on classifying $\mathbf{x}_i$,  $-a_iH_t(\mathbf{x}_i)$ will be greater than zero and thus, $y_t^i=1$. For the samples that are classified correctly, $-a_i H_t(\mathbf{x}_i)\le0$ and thus, $0\le y_t^i \le 1$. That is, $N\epsilon_t =\text{number of wrongly classified samples at round $t$}\le \sum_{i=1}^Ny_t^i=||\mathbf{y}_t||_1$ .

We are now ready to proceed with the proof of Theorem 5. Let $\mathbf{w}^*=[w_1^*,\cdots\!,w_N^*]^{\top}$ to be a vector where $w^*_i=1$ if $f(\mathbf{x}_i)\neq a_i$, and 0 otherwise. Similar to the proof of the lazy update, we are going to bound the $\sum_{t=1}^T (\mathbf{w}^*\!-\mathbf{y}_{t})^{\top}\eta_t\mathbf{d}_{t}$.

\begin{align}
\label{ap_t53}
(\mathbf{w}^*\!-\mathbf{y}_{t})^{\top}\eta_t&\mathbf{d}_{t}\!=(\mathbf{y}_{t+1}-\mathbf{y}_{t})^{\top}\big(\nabla\mathcal{R}(\mathbf{z}_{t+1})-\nabla\mathcal{R}(\mathbf{z}_{t})\big)\nonumber\\ &+(\mathbf{z}_{t+1}-\mathbf{y}_{t+1})^{\top}\big(\nabla\mathcal{R}(\mathbf{z}_{t+1})-\nabla\mathcal{R}(\mathbf{z}_{t})\big)
 +(\mathbf{w}^*-\mathbf{z}_{t+1})^{\top}\big(\nabla\mathcal{R}(\mathbf{z}_{t+1})-\nabla\mathcal{R}(\mathbf{z}_{t})\big)\nonumber\\
&\le  \frac{1}{2N}||\mathbf{y}_{t+1}-\mathbf{y}_t||^2+\frac{N}{2}\eta_t^2||\mathbf{d}_t||^2_*+B_{\mathcal{R}}(\mathbf{y}_{t+1},\mathbf{z}_{t+1})-B_{\mathcal{R}}(\mathbf{y}_{t+1},\mathbf{z}_{t})+B_{\mathcal{R}}(\mathbf{z}_{t+1},\mathbf{z}_{t})\nonumber\\
&-B_{\mathcal{R}}(\mathbf{w}^*,\mathbf{z}_{t+1})+B_{\mathcal{R}}(\mathbf{w}^*,\mathbf{z}_{t})-B_{\mathcal{R}}(\mathbf{z}_{t+1},\mathbf{z}_{t})\nonumber\\
&\le \frac{1}{2N}||\mathbf{y}_{t+1}-\mathbf{y}_t||^2+\frac{N}{2}\eta_t^2||\mathbf{d}_t||^2_*-B_{\mathcal{R}}(\mathbf{y}_{t+1},\mathbf{y}_{t})\nonumber\\
&+B_{\mathcal{R}}(\mathbf{y}_{t+1},\mathbf{z}_{t+1})-B_{\mathcal{R}}(\mathbf{y}_{t},\mathbf{z}_{t})-B_{\mathcal{R}}(\mathbf{w}^*,\mathbf{z}_{t+1})+B_{\mathcal{R}}(\mathbf{w}^*,\mathbf{z}_{t})
\end{align}
where the first inequality follows from applying Lemma 3 to the first term and Lemma 2 to the rest of the terms and the second inequality is the result of applying the exact version of Lemma 1 to $B_{\mathcal{R}}(\mathbf{y}_{t+1},\mathbf{z}_{t})$. Moreover, according to inequality \eref{ap_t5_1} $B_{\mathcal{R}}(\mathbf{y}_{t+1},\mathbf{y}_{t})-\!\frac{1}{2N}||\mathbf{y}_{t+1}-\mathbf{y}_t||^2\ge0$ and hence these terms can be ignored in \eref{ap_t53}. Summing up the inequality \eref{ap_t53} from $t=1$ to $T$, yields:
\begin{align} 
\label{ap_t54}
  -B_{\mathcal{R}}(\mathbf{w}^*,\mathbf{z}_{1})\le \sum_{t=1}^T \frac{N}{2}\eta_t^2  - \sum_{t=1}^T \eta_t\gamma_t ||\mathbf{y}_t||_1
\end{align}
It is important to remark that $||\mathbf{y}_t||_1$ appearing in the last term is due to the fact that $\mathbf{w}_t= \frac{\mathbf{y}_t}{||\mathbf{y}_t||_1}$ and thus, $\mathbf{y}_{t}^{\top}\eta_t\mathbf{d}_{t} = \mathbf{w}_{t}^{\top}\eta_t\mathbf{d}_{t} ||\mathbf{y}_t||_1 = \eta_t\gamma_t  ||\mathbf{y}_t||_1 $.

Now, by replacing $\eta_t=\epsilon_t\gamma_t$ in the above equation and noting that $B_{\mathcal{R}}(\mathbf{w}^*,\mathbf{z}_{1}) = N-N\epsilon$, we get:
\begin{align} 
\label{ap_t54}
  -N(1-\epsilon) \le \sum_{t=1}^T \frac{N}{2}\epsilon_t^2\gamma_t^2 - \sum_{t=1}^T \epsilon_t\gamma_t^2 ||\mathbf{y}_t||_1
\end{align}
From Lemma 6, it is evident that  $||\mathbf{y}_t||_1\ge N\epsilon_t$. Moreover, since $\epsilon \le \epsilon_t$, it can be replaced by $\epsilon$, as well (though very pessimistic). As usuall, $\gamma_t$ is also replaced with the min edge, denoted by $\gamma$. Applying these lower bounds to the equation \eref{ap_t54}, yields
\begin{align} 
\label{ap_t55}
\epsilon^2 \le  \frac{2(1-\epsilon)}{T\gamma^2}\le  \frac{1}{T\gamma^2}
\end{align}
which indicates that the proposed version of MadaBoost needs at most $O(\frac{1}{\mathlarger{\epsilon^2\gamma^2}})$ iterations to converge.

\end{document}